\documentclass[11pt,a4paper, table]{article}
\usepackage[hyperref]{emnlp2020}
\usepackage{times}
\usepackage{latexsym}

\usepackage{microtype}

\aclfinalcopy % Uncomment this line for the final submission
\def\aclpaperid{1135} %  Enter the acl Paper ID here

\usepackage{multirow}
\usepackage{graphicx}
\usepackage{booktabs}
\usepackage{float}
\usepackage{amsmath}
\usepackage{enumitem}
\usepackage{nidanfloat}

\usepackage{titlesec}
\title{What time is it? Temporal Analysis of Novels}

\author{Allen Kim, Charuta Pethe, Steven Skiena \\
  Department of Computer Science, \\ Stony Brook University, NY, USA \\
  {\texttt{\{allekim,cpethe,skiena\}@cs.stonybrook.edu}}}

\date{}

\begin{document}
\maketitle
\begin{abstract}
Recognizing the flow of time in a story is a crucial aspect of understanding it.  Prior work related to time has primarily focused on identifying temporal expressions or relative sequencing of events, but here we propose computationally annotating each line of a book with wall clock times, even in the absence of explicit time-descriptive phrases. To do so, we construct a data set of hourly time phrases from 52,183 fictional books. We then construct a time-of-day classification model that achieves an average error of 2.27 hours. Furthermore, we show that by analyzing a book in whole using dynamic programming of breakpoints, we can roughly partition a book into segments that each correspond to a particular time-of-day. This approach improves upon baselines by over two hours. Finally, we apply our model to a corpus of literature categorized by different periods in history, to show interesting trends of hourly activity throughout the past. Among several observations we find that the fraction of events taking place past 10 P.M jumps past 1880 - coincident with the advent of the electric light bulb and city lights.
\end{abstract}

\section{Introduction}
\label{sec:intro}

The flow of time is an indispensable guide for our actions, and provides a framework in which to see a logical progression of events.
Just as in real life, the clock provides the background against which literary works play out: when characters wake, eat, and act.  In most works of fiction, the events of the story take place during recognizable time periods over the course of the day.   Recognizing a story's flow through time is essential to understanding the text.

In this paper, we try to capture the flow of time through novels by attempting to recognize what time of day each event in the story takes place at. 

\begin{figure*}
\centering
\includegraphics[width=\textwidth]{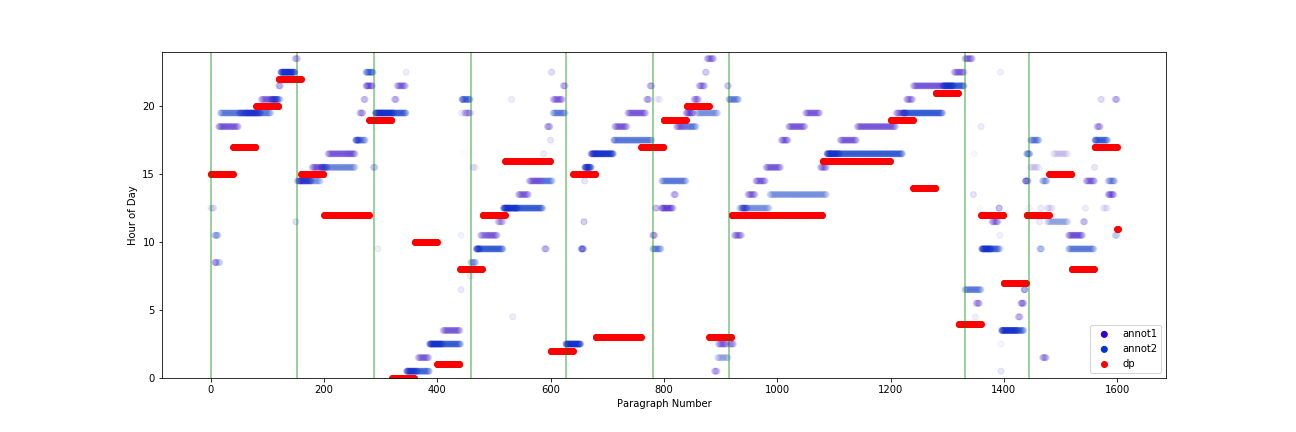}
\caption{Comparing annotator times (in blue) versus algorithmically generated times for {\em The Great Gatsby}, shown in red. Blue colored dots represent the common agreed upon points of time. Green lines represent start of chapters. (A half-hour offset between the annotator and the model prediction to eliminate direct overlaps).}
\label{fig:gatsby}
\end{figure*}

As our motivating example, we use {\em ``The Great Gatsby''} \cite{fitzgerald_2004}, a short novel with a familiar plot that can be analyzed with our techniques. Figure \ref{fig:gatsby} presents the work of two human annotators, independently making their best guesses of the clock time at every paragraph of the book. The x-axis of Figure \ref{fig:gatsby} represents the full text of the book as enumerated by paragraph numbers, while the y-axis represents the 24 hours of the day.  The times identified by the annotators are shown in blue, while our model's
time predictions are shown in red. 

We first note that there is general but not perfect agreement between the annotators, with an average disagreement of 1.85 hours.
There is also a strong general agreement between the model and the annotators, with an average absolute error of 2.62 hours, computed by taking the minimum time difference from either of the annotators. 

\begin{figure}
\centering
\includegraphics[width=7.5cm]{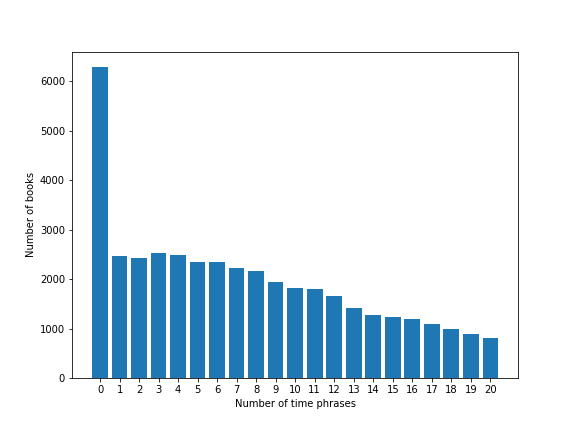}
\caption{Number of books with explicit time references out of 52,183 books. 12.04\% of these books contain no clock times within them.}
\label{fig:timeperbook}
\end{figure}

Although human readers are generally able to track the flow of texts, this task is more difficult than may initially be supposed -- because there are surprisingly few actual times reported in most books.
Figure \ref{fig:timeperbook} shows how many explicit hourly time phrases appear in our dataset of 52,183 novels. About 6,000 of these books contain \emph{zero} time references to any hour.
Among the books that do contain clock times, most of them contain fewer than twenty references. Events are often described in a neutral manner that does not signal much about exactly when it is taking place.

This scarcity of time references presents a big challenge in developing good models and in interpreting the evidence associated with a particular text.
Because of the lack of explicit references, some notion of likelihood must be part of the model. 
For this reason, we model time as a probability distribution of what hour it likely is at this juncture. 

Our contributions\footnote{Code and links to dataset can be found at \url{https://github.com/allenkim/what-time-is-it}} in this paper include:

\begin{itemize}
 
    \item \textbf{Literary Time Reference Dataset} -- We build a clean data resource containing all explicit time references in a dataset of 52,183 novels whose full text is available via Project Gutenberg \cite{gutenberg} and the HathiTrust Digital Library\footnote{\url{www.hathitrust.org}}.  The times extracted via regular expressions generally do not include AM or PM designations, so we build models to predict the AM/PM label for a window of text with the best model achieving an accuracy of 86.3\% at this task.
    
    \item \textbf{Models for Local Time Prediction} -- We develop three language-based models to forecast the time (on the military hour scale from 0 to 23) from local text windows.  This task is distinguished from typical regression problems in that time is periodic: thus the difference between 23:00 and 1:00 is the same as that between 10:30 and 12:30.
    We treat this task as a 24-class classification problem, with our strongest model (based on BERT) achieving an average error of 2.27 hours.
 
    \item \textbf{Global Time-Flow Analysis in Novels} -- Predictions of time signals from local text features are doomed to be of bounded accuracy because, as previously detailed, books contain relatively rare explicit time references.   Significant episodes typically require several pages to present, so we anticipate times to hold constant through large segments of text, and then proceed in a forward direction.
    
    We define an optimization criteria to partition texts into coherent time windows, and provide an efficient dynamic programming algorithm, which reduces the average absolute prediction error by over an hour against baselines.
    
    \item \textbf{Historical Trends in Hourly Activities} -- By analyzing times extracted from our corpus of novels and the lifespan of its authors, we can identify how waking periods and peak times of activity have changed over the past two hundred years.  In particular, we demonstrate that the fraction of events in novels taking place after 10 PM has grown steadily since 1880 -- a tribute to the power of the electric light.   Characters rose with the sun far more often in the agrarian society of old.   Contemporary characters spend more time at lunch and less at dinner than their forebearers.
    
\end{itemize}

Our paper is organized as follows.  We first discuss related works in temporal analysis and NLP for fiction (Section \ref{sec:related}). We then describe our data collection process, specifically how we extracted temporal expressions along with some analysis of the phrases (Section \ref{sec:dataset}). We then describe how the time-of-day models were constructed and their evaluations (Section \ref{sec:time}). With these models, we show our book-length prediction algorithms and their corresponding metrics as well (Section \ref{sec:book}). Finally, we present a trend of time activity over history based on our book-length prediction algorithm (Section \ref{sec:gutenberg}).

\section{Related Works}
\label{sec:related}

Work related to temporal analysis stems back to foundations in logic, which defined time in the context of sets and relations \cite{bruce1972model, allen1983maintaining}. Less formally, we associate time with events \cite{setzer2000annotating} and indeed most recent work has similarly approached understanding time in the context of events.

There has been much work done on temporal event analysis, primarily in the areas of identifying time phrases as well as extracting temporal relations between them \cite{pustejovsky2005temporal, bramsen2006finding, chambers2007classifying, bethard2007cu, lapata2006learning}. Standards have been set up to properly categorize what kind of phrases are considered to be ``time phrases'' \cite{pustejovsky2003timeml}, and methods using regular expressions \cite{mani2000robust, strotgen2010heideltime} as well as machine learning approaches \cite{mani2006machine, min2007lcc, uzzaman2010event} have been used to parse these expressions. Looking more broadly, there has also been much work done in understanding a document's time dimension, which can involve determining when the document was created/published \cite{de2005temporal, garcia2011written} or determining the time in which the contents of the story is focused on \cite{jatowt2013estimating} or analyzing dates in German literature \cite{fischer2015does}.

Not only are we interested in just time, we are interested in time in the context of literature. There is a vast field of research in analyzing literature such as sentiment analysis of plot \cite{alm2005emotional, mohammad2013once, elsner2012character, reagan2016emotional, jockers2015revealing}. Since time is inherently connected to events, we also refer to literature in parsing literary events within books as well \cite{ahn2006stages, liao2010using, li2013joint, feng2018language, sims2019literary}. Within literature, we are also interested in the activities of humans over history. For example, electric lighting began to become more popularized in the form of lightbulbs and city lights in the 1880s \cite{dilaura2008brief}, and with electric lighting, people can more readily be active during the night. Prior to then, we are inclined to believe that people were not as active late at night.

\section{Dataset Preparation}
\label{sec:dataset}

 We initially started with a dataset of 10,489 Gutenberg books as well as 97,772 HathiTrust books, which were filtered to be English fiction \cite{underwoodhathilit}, but these contained numerous duplicates of the same book as well as duplicates between the Gutenberg dataset. These were deduplicated based on title and author similarity for a resulting count of 52,183 unique books over both sets. For the sake of completeness, we present results in this paper that use Gutenberg and HathiTrust books independently.
 
 Regarding the format of the data, the Gutenberg books are provided as a raw text files and were cleaned to strip headers and front matter. The HathiTrust books were provided as a folder of text files representing pages. These were preprocessed to strip headers using the HathiTrust Research Center RunningHeaders tool \footnote{\url{https://github.com/htrc/HTRC-Tools-RunningHeaders-Python}} as well as to separate out the body of the book from its front and back matter \cite{mcconnaughey2017labeled}. We also performed further preprocessing to split them into paragraphs, sentences, and tokens. Among the preprocessing tasks, the most important task was to annotate which tokens belong to time phrases. This was done using SUTime library \cite{chang2012sutime} to tag time terms, which uses a regular expression based approach. In particular, we focused on times that pinpointed hours within a day such as ``two o'clock'' or ``noon''. The library provides the terms as well as the time it can be translated to.

\begin{table}
\def\arraystretch{1.25}
\centering
\rowcolors{3}{gray!15!}{}
\begin{tabular}{ c*{2}{rrrr}}
\\ \hline
\multirow{3}{*}{} & \multicolumn{2}{c}{\textbf{Gutenberg}} & %
    \multicolumn{2}{c}{\textbf{HathiTrust}}\\
\textbf{Hour} & \textbf{A.M} & \textbf{P.M} & \textbf{A.M} & \textbf{P.M} \\\midrule
0    & 21,810 & 21,646 & 123,214 & 121,649 \\
1    & 1,038   & 315   & 29,582 & 1,696\\
2    & 1,337   & 661   & 7,330 & 3,582\\
3    & 1,139   & 965   & 6,060 & 4,867\\
4    & 911   & 1032   & 5,133 & 5,150\\
5    & 645   & 720   & 4,053 & 3,693\\
6    & 754   & 562   & 4,316 & 3,125\\
7    & 531   & 495   & 3,421 & 2,602\\
8    & 698   & 596   & 3,950 & 3,401\\
9    & 657   & 587   & 3,795 & 3,468\\
10   & 745   & 661   & 3,952 & 3,897\\
11   & 410   & 460   & 2,055 & 2,737\\ \bottomrule
\end{tabular}%
\caption{Number of time examples by hour.  The most frequent explicit time references are to noon and midnight.}
\label{tab:numtimes}
\end{table}

Table \ref{tab:numtimes} shows the number of time tokens for every hour. The first two columns show the number of time references that can be determined down to an hour out of the 24 hour clock for Gutenberg while the latter two are for HathiTrust. However, these are only for known time references; there also exists ambiguous references. Time phrases can be disambiguated with words such as ``AM'' or ``PM'', or if the time is connected to a prepositional phrase such as ``three in the afternoon''. However, it is more often the case that time phrases are used without explicit labels. References to a certain time like ``eight o'clock'' usually do not come with an ``AM'' or ``PM'' tag as it is often inferred from context. Note that we especially have many samples for hour 0 - midnight and noon - simply because those are more commonly used phrases as opposed to prepositional phrases such as ``five o'clock at night''. We also note that there are abnormally large amounts of examples for 1 A.M. in the HathiTrust dataset. We found that this was largely due to OCR errors that transcribed ``I am'' to ``1 am''.

\subsection{Associating Words with Hours of the Day}

Given the references in each hour, we now want to examine words that are over-represented in different time periods. For each time reference, we take a window including three sentences before and after the reference, concatenate all of the windows referring to the same hour, and tokenize and count the words among them. This effectively creates 24 bag of words, one for each hour in the day. Since time periods with a difference of one hour are practically similar in the context in which they can appear, we also merge neighboring hour bag of words with each other.

We now want to determine how closely related a word is to a particular hour. To do so, we define a scoring function that takes in a word and hour, and outputs a score between 0 and 1 representing how closely related the word is to that hour. As examples, we would expect the word ``lunch'' and the hour 12 to have a high score and the word ``dinner'' and the hour ``8'' to have a low score.
\begin{table}
\def\arraystretch{1.25}
\centering
\rowcolors{3}{gray!15!}{}
\begin{tabular}{cr}
\\ \hline
\textbf{word} & \textbf{top three hours} \\\midrule
breakfast  & 7, 8, 6 \\
bright  & 10, 11, 12 \\
sun  & 12, 13, 11 \\
lunch  & 12, 13, 11 \\
park  & 15, 16, 11\\ 
dinner  & 18, 19, 13\\ 
dark  & 23, 0, 1\\ 
moon  & 23, 0, 1\\ 
\bottomrule
\end{tabular}%
\caption{Top three hours for select feature words, consistent with common experience.}
\label{tab:hourwords}
\end{table}
\paragraph{Scoring Description} Intuitively, we want to measure how surprised we are when observing the frequency of word in the context of a hourly phrase versus the frequency of the word in any standard text. For a given word and hour, we look at the frequency of the word within the hour's bag of words and compare it to the standard normalized frequency of the word in all of our dataset. We model the occurrence of the word as a geometric distribution with probability equal to the standard normalized frequency, and thus, we can score each word by using their frequencies within the bag of words and using a binomial cumulative distribution function to find a corresponding likelihood. 

Formally, for a given word $w$ and hour $h$, let $k$ be the number of occurrences of $w$ in the bag of words of $h$, $p[w]$ be the probability of $w$ appearing in a text -- computed by taking the normalized frequency of $w$ in all of the text -- and $N_h$ be the total number of words in the bag of words of $h$. We then define the scoring function $s(w,h)$ as:
$$s(w,h) = \sum_{i=0}^{k} \binom{N_h}{i}p[w]^i(1-p[w])^{N_h-i}$$

With this scoring function, for any given word, we can rank the top hours it is associated with. We show the top three hours for several common words in Table \ref{tab:hourwords}. We considered words with common associated times with them such as eating breakfast in the morning, lunch at noon, and dinner at night. We see that there is a general agreement between the top hours for the given words and the times an average person would associate them with.

\section{Time Prediction from Text}
\label{sec:time}

For a given window of sentences, we seek to predict the hour it is most likely taking place in. To construct models for this task, we use the time phrases from Section \ref{sec:dataset}, but we note that there exists far more unlabeled time phrases. Given the limited amount of labeled data, we first want to augment our dataset by labeling the unlabeled data as well. Thus, we have a two step approach:

\begin{enumerate}
    \item Build a model to resolve ambiguous time terms (AM versus PM) and label the unlabeled data.
    \item Train a model for time of day prediction by hour using the augmented dataset.
\end{enumerate}

\subsection{Resolving Ambiguity in Time Terms}

We first remark that the majority of time terms do not have AM or PM tags. There are approximately three times as many unlabeled time terms than labeled ones, excluding noon and midnight. In order to get rid of this ambiguity, we first label this data with AM/PM tags.

Our problem is as follows: Given a reference to a phrase representing some hour of the day as well as the words in the context around it, determine whether the time the phrase is referring to is ``A.M'' or ``P.M''. Intuitively, this problem requires 12 different models, one for each hour from 0 to 11. We require separate models for each hour since neighboring hours can have similar words but different labels. For example, ``11 A.M'' can have words quite similar to ``12 P.M'', but their labels are clearly different. Additionally, we use a window that spans three sentences before the time phrase and three sentences afterwards. We found that empirically, the predictive power of the model did not significantly improve past this window size.

We consider three main models along with a baseline and an ensemble of the three models.

\begin{itemize}
    \item \textbf{Baseline:} We use the majority class for each hour as the default prediction.
    \item \textbf{Naive Bayes (NB):} We convert the window of sentences to a binary bag of word representation using StanfordNLP \cite{qi2019universal} for every hour and train Naive Bayes classifier for prediction.
    \item \textbf{LSTM:} We represent the window of sentences as vectors using GloVe \cite{pennington2014glove}. We use the 6B tokens, 400K vocab, uncased, 100d pre-trained word vectors to convert windows to sequences of vectors that were then used to train an LSTM.
    \item \textbf{BERT:} We tokenize the windows using the BERT tokenizer and fine-tune the pre-trained 12-layer, 768-hidden, 12-heads, 110M parameters uncased BERT model \cite{devlin2018bert}.
\end{itemize}

\paragraph{Experimental Details.}  For each of these model types, we construct twelve classifiers (one for each hour up to twelve). We also split the training and testing set in a 70-30\% split and further split the training set in a 90-10\% split for validation. Additionally, we take advantage of the fact that the windows of words around one hour are quite similar to the hours near it. We can imagine that replacing ``1 P.M'' with ``2 P.M'' in a window will have minimal impact. Thus, we take the neighboring hours training set as well when training for each hour.

All models were run on a compute server with 2.30 GHz CPU and TeslaV100 GPU. No hyperparameter tuning was done on any models; default values were run for all models. The average training time of all the neural models were within several hours. This is true for future experimentation as well.

One point to note is the necessity of masking the time phrase that the window was based on. Words such as ``AM/PM'' or ``in the morning/night'' provided the temporal cues to parse the time phrase in the first place, and any decent model with access to these words will perform with unrealistic accuracy. These features do not exist in the unlabeled training set, to ensure the models learn to identify the proper AM/PM label without cheating. Thus, we replace all the time phrases with the same special token. In some models such as BERT, we use this while tokenizing. In other models such as GloVe + LSTM, we create a new random vector to represent this token.

\begin{table}
\def\arraystretch{1.25}
\centering
\rowcolors{4}{gray!15!}{}
\begin{tabular}{*{5}{c}}
\\ \hline
\textbf{classifier} & \textbf{type} & \textbf{acc} & \textbf{am f1} & \textbf{pm f1} \\\midrule
\multicolumn{5}{l}{\underline{\textbf{Gutenberg}}}  \\
\cellcolor{white} & mic & 0.520 & 0.306 & 0.633 \\
     \cellcolor{white}\multirow{-2}{*}{Baseline} & mac & 0.573 & 0.501 & 0.223 \\\addlinespace
\cellcolor{white} & mic & 0.704 & 0.715 & 0.692 \\
\cellcolor{white}\multirow{-2}{*}{NB}  & mac & \textbf{0.671} & 0.688 & \textbf{0.603} \\\addlinespace
\cellcolor{white} & mic & 0.713 & 0.728 & 0.699 \\
\cellcolor{white}\multirow{-2}{*}{LSTM}  & mac & 0.585 & 0.622 & 0.510 \\\addlinespace
\cellcolor{white} & mic & \textbf{0.793} & \textbf{0.800} & \textbf{0.785} \\
\cellcolor{white}\multirow{-2}{*}{BERT} & mac & 0.665 & \textbf{0.695} & 0.601 \\ \hline
\multicolumn{5}{l}{\underline{\textbf{HathiTrust}}} \\
\cellcolor{white}\multirow{2}{*}{Baseline} & mic & 0.621 & 0.738 & 0.312 \\
      & mac & 0.576 & 0.554 & 0.171 \\\addlinespace
\cellcolor{white}\multirow{2}{*}{NB}  & mic & 0.739 & 0.786 & 0.665 \\
      & mac & 0.729 & 0.748 & 0.666 \\\addlinespace
\cellcolor{white}\multirow{2}{*}{LSTM} & mic & 0.766 & 0.810 & 0.697 \\
      & mac & 0.723 & 0.728 & 0.674 \\\addlinespace
\cellcolor{white}\multirow{2}{*}{BERT} & mic & \textbf{0.863} & \textbf{0.889} & \textbf{0.821} \\
      & mac & \textbf{0.837} & \textbf{0.847} & \textbf{0.804} \\
\bottomrule
\end{tabular}%
 \caption{AM - PM Prediction Results for Gutenberg / HathiTrust}
 \label{tab:ampm}
\end{table}

\paragraph{Results.} The results are shown in Table \ref{tab:ampm} with the metrics of accuracy and F1 scores for each class. We include results when running these models purely on Gutenberg data as well as the results when running these models with the HathiTrust data. We see clear improvement across the board, especially for BERT with the extra data. The macro metrics are the averaged values over all 12 models while the micro metrics are the values over all the test examples over all the models. The results show that this is a challenging task given the limited amount of data we have. Analyzing the dataset shows that many windows that contain a time reference can be sensible with either A.M or P.M, so it is not easy to disambiguate mentions of time in generic dialogue. However, all our models substantially outperform the baseline.  In the end, we use the winning BERT model to label our unlabeled data for training.

Since we are imputing our data with computer-generated labels, we compare its output to human annotators to test how reliable it is. We manually annotated 1200 instances of AM/PM windows --- 100 examples for each hour pair --- and compared it to our model's output to see the agreement. In cases where the label can be ambiguous, the annotators made their best intuitive guess. Figure \ref{tab:annotampm} shows the agreement between the annotators and our model, where the agreement is measured in number of matching predictions divided by the total number.

\begin{table}
\def\arraystretch{1.25}
\centering
\rowcolors{3}{gray!15!}{}
\begin{tabular}{cr|cr}
\\ \hline
\textbf{hour} & \textbf{agreement} & \textbf{hour} & \textbf{agreement}\\\midrule
0    & 0.79  & 6 & 0.76 \\
1    & 0.69 & 7    & 0.76 \\
2    & 0.79 & 8    & 0.80 \\
3    & 0.82 & 9    & 0.69 \\
4    & 0.75 & 10   & 0.71 \\
5    & 0.80 & 11   & 0.77 \\
\bottomrule
\multicolumn{4}{c}{Mean Agreement = 0.761} \\ 
\end{tabular}%
\caption{Agreement between Annotators and Model for AM/PM prediction
}
\label{tab:annotampm}
\end{table}

We see that the model performs respectably with an overall average of 76\% accuracy. We comment that the human annotations contain some anomalies due to linguistic changes such as ``dinner'' being eaten as lunchtime and ``supper'' being the canonical name for a later meal, but these anomalies were relatively minor.

\subsection{Time of Day Prediction}

Given the words in the context of the time phrase, we now predict the most likely time of day. We treat this problem as a 24-class classification problem, where each class is defined to be an hour of the day. We again consider the same three models as in the AM-PM models: bag of words with Naive Bayes, GloVe with LSTM, and BERT fine-tuning, but with 24-class outputs as opposed to binary. 

\paragraph{Results.} The results are shown in Table \ref{tab:24hour} and \ref{tab:24houravg}. The models shown in these results were trained on exclusively HathiTrust books. We also show the results when purely trained on Gutenberg books as well. We note that error is measured in number of hours. Thus, the worst possible error is 12 hours on a 24 hour clock. A baseline model with random guessing would have an expected error of 6 hours.

\begin{table}
\def\arraystretch{1.25}
\centering
\rowcolors{3}{gray!15!}{}
\begin{tabular}{crrr}
\\ \hline
\textbf{hour} & \textbf{NB} & \textbf{LSTM} & \textbf{BERT} \\\midrule
0    & 2.92 & 2.56 & \textbf{1.69} \\ 
1    & 2.07 & 1.56 & \textbf{1.05} \\
2    & 4.24 & 3.28 & \textbf{2.57} \\
3    & 5.21 & 3.54 & \textbf{2.71} \\
4    & 5.75 & 3.51 & \textbf{2.57} \\
5    & 5.94 & 3.87 & \textbf{2.66} \\
6    & 5.68 & 3.85 & \textbf{2.88} \\
7    & 5.72 & 4.01 & \textbf{2.85} \\
8    & 5.05 & 3.89 & \textbf{2.55} \\
9    & 5.04 & 3.81 & \textbf{2.58} \\
10   & 5.11 & 4.13 & \textbf{2.87} \\
11   & 5.10 & 3.63 & \textbf{2.27} \\
12   & 5.21 & 3.46 & \textbf{1.73} \\
13   & 4.98 & 3.29 & \textbf{1.88} \\
14   & 4.38 & 3.41 & \textbf{2.18} \\
15   & 3.86 & 3.32 & \textbf{2.12} \\
16   & 3.57 & 3.32 & \textbf{2.00} \\
17   & 3.57 & 3.19 & \textbf{2.23} \\
18   & 3.86 & 3.51 & \textbf{2.31} \\
19   & 4.22 & 3.29 & \textbf{2.48} \\
20   & 3.79 & 3.29 & \textbf{2.10} \\
21   & 3.42 & 3.20 & \textbf{2.01} \\
22   & 3.25 & 3.17 & \textbf{2.19} \\
23   & 3.17 & 2.61 & \textbf{2.17} \\ \bottomrule
\end{tabular}
\caption{Time-of-day prediction error by hour for HathiTrust books
}
\label{tab:24hour}
\end{table}

\begin{table}
\def\arraystretch{1.25}
\centering
\rowcolors{3}{gray!15!}{}
\begin{tabular}{crrr}
\\ \hline
\textbf{hour} & \textbf{NB} & \textbf{LSTM} & \textbf{BERT} \\\midrule
Gutenberg & 4.69 & 4.72 & \textbf{4.09} \\
HathiTrust  & 4.38 & 3.36 & \textbf{2.28}
 \\ \bottomrule
\end{tabular}
\caption{Average time-of-day prediction error for Gutenberg and HathiTrust books
}
\label{tab:24houravg}
\end{table}

We see that the BERT model performs the best with an error of 2.28 hours while Naive Bayes performs the worst with an error of 4.38 hours. We clearly see that this problem is heavily influenced by the amount of data available. We see that by simply adding more data to the LSTM and BERT models, the average error improves significantly and unsurprisingly, the naive Bayes model only improves slightly.

\section{Book-length Time Prediction}
\label{sec:book}

Given a model that can predict the time of day for a single window of sentences, we now consider predicting the time of day over an entire book -- constructing a time flow through the book. The simplest idea is to partition the book into windows that fit into the model and independently predict an hour for each window using the model. However, this will have very poor performance since many windows will consist of sentences that have no bearing to time and in these cases, the model will output an arbitrary time that will not fit with its surroundings. To resolve this, we consider the problem of optimally partitioning the windows into larger segments corresponding to particular hours.

More formally, given a sequence of sentence windows $s_1, s_2, \ldots, s_n$ and the number of segments, parameterized as $k$, our goal is to generate the most likely list of indices $i_1, i_2, \ldots, i_k$ that represent the start of each segment, and a list of hours $h_1, h_2, \ldots, h_k$ that represents the corresponding hour assigned to each segment.

\subsection{Generating Probability Distributions}

For every window, we now want to generate a probability distribution over the 24 hours. We present two different means of acquiring these probabilities and in the end, we combine these two probability distributions.

\paragraph{Model Probabilities.}
The first approach applies our BERT time of day model from Section \ref{sec:time}. As discussed in the introductory paragraph of this section, we can simply run our model on each window of text and min-max normalize the scores to get probabilities. We additionally smooth the probability by averaging the probabilities with their neighboring hours since we expect neighboring hour classes to be similar to each other. With this, we now have a probability distribution over 24 hours for each window using our model. However, one limiting feature of just using our model is the fact that our model was trained with core time phrases removed from its training. Recall that phrases such as ``eight o'clock at night'' were masked entirely to prevent skewing the testing procedure.

\paragraph{Tag-based Probabilities.} To make use of these crucial time phrases, we consider another probability distribution based on key time phrases that are also annotated by SUTime: ``morning'', ``afternoon'', ``evening'', and ``night''. For each tag, we define a probability over the standard hours in which they refer to (morning: 6-11, afternoon: 12-16, evening: 17-20, night: 21-5). For any window containing these tags, we define a uniform probability over the hours the tag refers to with zeros for other hours. This defines a probability distribution for windows containing these tags. For windows not containing any time tags, we simply let it equal the probability of the previous known tag.

\paragraph{Merging.} We average our model probabilities with the tag-based probabilities to get our final probability distribution for every window.

\subsection{Optimal Partitioning of Probabilities}

Given the probability distribution for each window, we now want to find the optimal partitioning of the windows. Recall that in the formulation of this problem, one of the required parameters is the number of partitions we want to make among the windows of text. We consider this number as a parameter we can control. If we allow too many segments, then the model will probably overfit to noisy windows whereas if the number of segments is too small, then the times will not be accurate to the book. For our experiments, we approximate the ratio of the number of windows to number of partitions in a book to be approximately eight, which is about 55 sentences on average per partition. We saw that this worked well empirically with several sample texts such as \emph{The Great Gatsby} and maintained a good balance between not overfitting and getting sensible results.

To determine the location of partition breaks, we present a baseline and two methods.

\begin{itemize}

    \item \textbf{Baseline:} We assign every window to be the same constant hour - we choose noon in particular since that is the middle of the day.
    \item \textbf{Max Hour:} We first partition the windows into equal sizes. Each partition now contains a series of probability distributions. For each partition, we take the sum of the probabilities across each window and assigns to each partition the hour that corresponds to the maximal sum in the summed partition probabilities.
    \item \textbf{Dynamic Programming (DP):} The dynamic programming  takes in the number of text windows and number of partitions as parameters and optimizes the size of the segments to maximize the alignment of each section with its underlying probability. We define the DP recurrence relation to score the alignment $f(n,k)$, where $n$ is the number of windows and $k$ is the number of breaks, as:
% \begin{align*}
$$\max_{i \in [1,n-k]} \bigg( f(n-i,k-1) + \max_{h\in [0,23]} \sum_{j=n-i}^n p_h[j] \bigg)$$
% \end{align*}
where we define $p$ as the array of array of 24 probabilities for every window, and thus, $p[j]$ can be described as an array of 24 probabilities for window $j$. By taking the max, the DP prioritizes the hour with maximal probability sum over the length of the partition.
\end{itemize}

\subsection{Evaluation}

To evaluate our methods, we construct ground truth for the books in our dataset. While the example with ``The Great Gatsby'' was manually annotated, we have no annotations of times for our book dataset. Thus, we approximate the ground truth by considering books that contain time references to specific hours of the day and annotate the text window containing that phrase to be that hour. Additionally, to raise the quality of the test set, we only consider books with multiple time references that include references beyond just noon and midnight. We note that our ``ground truth'' is not correct in many circumstances such as when a specific time is referred to in dialogue referencing some point in the past or future, but suffices to show general trends in our results.

\begin{table}
\def\arraystretch{1.25}
\centering
\rowcolors{3}{gray!15!}{}
\begin{tabular}{crr}
\\ \hline
\textbf{classifer} & \textbf{error} \\\midrule
Noon Baseline & 6.215   \\ 
Max Hour   & 4.250 \\
DP   & 4.232   \\ 
\bottomrule
\end{tabular}%
\caption{Book Time Prediction Results}
\label{tab:booktime}
\end{table}

\paragraph{Results.} Our results for average error in hours are shown in Table \ref{tab:booktime}. Even with the low quality of labels in the test set, we see that the local method of maximal hours over uniform segments as well as the dynamic programming method beat the noon baseline by about two hours. Overall, the dynamic programming method performs best, but the local maximization method performs admirably as well.

One might ask why the error is higher compared to the 24 hour model. This is due to the fact that while the model performs well on local windows that contain a time reference, the neighboring windows tend to give little signal about time and confuses which windows should be emphasized more than others. Quite often, the probabilities provided by the models do not fully represent confidence. For future work, it would be worth considering a different model that uses probabilistic labels as classes or a variant of a regression model.

\section{Historical Trends}
\label{sec:gutenberg}

In the end, we use our book-length time prediction through all the Gutenberg books in our dataset and found the distribution of times throughout each book. While the dataset does not have the publication date of the book in the metadata, we were able to access the authors and the years the author was alive. Thus, we created groupings of our data by time period based on the year of the author's birth. We group the authors by year of birth, separated every 20 years.

\begin{figure*}
\centering
\includegraphics[width=\textwidth]{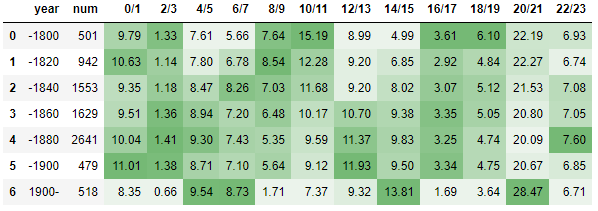}
\caption{Activity in Different Time Periods by Percentage - shaded by relative weight within each column}
\label{fig:periodactivity}
\end{figure*}

Figure \ref{fig:periodactivity} shows the results for six groupings of years. We consider all books up to 1800 to be one group, then every 20 years afterwards up to 1900, and all books from 1900 and beyond to be the final group. We first note the number of books in each grouping. Given copyright laws, most of the books we have access to are in the late 1800s as shown by the number column.

We see some interesting trends. For example, noon and afternoons (12 to 5 P.M) are referenced more as the periods pass. Additionally, the times around dinner (6 to 9 P.M.) are referenced much more in earlier books, but become less relevant in later periods. The emphasis tends to shift towards later at night (11 P.M to 1 A.M). We also see a decrease in emphasis in the morning from 8 to 9 A.M. Overall, the emphasis seems to shift towards the afternoon and late nights and away from mornings and evenings as we progress through the books historically, which can potentially be attributed to the rise in the emphasis on lunch times as well as the advent of electric lighting, along with a decreased emphasis on family dinners.

\section{Conclusions and Future Work}
\label{sec:conclusion}

We have constructed a dataset of time phrases to build models that can predict the most relevant hour of the day for a given text window. Our models are a good start, but we release the dataset to encourage others to improve on this task. We note that this dataset can be further cleaned by resolving OCR errors in the source text as well as improving upon the time extraction algorithm. More annotations of complete novels would permit better models and evaluation.

Full time annotations of novels additionally include the challenging task of distinguishing between narrator and recall time in discussing past events. We also seek to annotate information about dates and seasons. Future work includes applying time inference models to question answering and other NLP systems.

\section*{Acknowledgments}
We thank the reviewers for their feedback. This work was partially supported by NSF grants IIS-1926751,
IIS-1927227, and IIS-1546113.

\bibliographystyle{acl_natbib}
\bibliography{emnlp2020}

\end{document}